\documentclass[letterpaper, 10 pt, conference]{ieeeconf}  
\PassOptionsToPackage{table}{xcolor}

\IEEEoverridecommandlockouts                              

\usepackage{cite}
\usepackage{amsmath,amssymb,amsfonts}
\usepackage{dsfont}
\usepackage{balance}
\usepackage[font=footnotesize]{caption}
\usepackage{algorithmic}
\usepackage{graphicx}
\usepackage{textcomp}
\usepackage{xcolor}
\usepackage{booktabs} 
\usepackage{lipsum} 
\usepackage{hyperref}
\usepackage{booktabs} 
\usepackage{multirow} 
\usepackage{adjustbox} 
\usepackage{makecell}
\usepackage{balance}  
\usepackage{flushend}
\usepackage{colortbl}
\usepackage[table]{xcolor}
\usepackage{bm} 

\def\secref#1{Sec.~\ref{#1}}
\def\figref#1{Fig.~\ref{#1}}
\def\tabref#1{Tab.~\ref{#1}}
\def\eqref#1{Eq.~(\ref{#1})}

\definecolor{lightyellow}{RGB}{255, 255, 200}
\definecolor{upperblue}{RGB}{99, 112, 250}
\definecolor{bottomred}{RGB}{239, 99, 74}
\definecolor{bottomgray}{RGB}{116, 116, 116}

\usepackage{fancyhdr}
\fancypagestyle{arxivhdr}
{
   \fancyhf{}
   \setlength{\headheight}{15pt} 
\fancyfoot[C]{This paper has been accepted for publication at the 2025 IEEE/RSJ International Conference on Intelligent Robots and Systems (IROS)}
\fancyhead[C]{\footnotesize Please cite this paper as:\\
Wanmeng Li, Simone Mosco, Daniel Fusaro, and Alberto Pretto, "DPGLA: Bridging the Gap between Synthetic and Real Data for Unsupervised Domain Adaptation in 3D LiDAR Semantic Segmentation," 2025 IEEE/RSJ International Conference on Intelligent Robots and Systems (IROS)}
}

\title{\LARGE \bf
DPGLA: Bridging the Gap between Synthetic and Real Data for Unsupervised Domain Adaptation in 3D LiDAR Semantic Segmentation
}

\author{Wanmeng Li\hspace{6em}Simone Mosco\hspace{6em}Daniel Fusaro\hspace{6em}Alberto Pretto
\thanks{All authors are with the Intelligent Autonomous Systems Laboratory (IAS-LAB), Department of Information Engineering of the University of Padua, Italy. 
E-mail addresses are as follows: \texttt{\{liwanmeng, moscosimon, fusarodani, albertopretto\}@dei.unipd.it}
.
This project is partially supported by the China Scholarship Council.}
}

\begin{document}

\maketitle
\thispagestyle{empty}
\pagestyle{empty}
\thispagestyle{arxivhdr}

\begin{abstract}
Annotating real-world LiDAR point clouds for use in intelligent autonomous systems is costly. To overcome this limitation, self-training-based Unsupervised Domain Adaptation (UDA) has been widely used to improve point cloud semantic segmentation by leveraging synthetic point cloud data. However, we argue that existing methods do not effectively utilize unlabeled data, as they either rely on predefined or fixed confidence thresholds, resulting in suboptimal performance. In this paper, we propose a Dynamic Pseudo-Label Filtering (DPLF) scheme to enhance real data utilization in point cloud UDA semantic segmentation. Additionally, we design a simple and efficient Prior-Guided Data Augmentation Pipeline (PG-DAP) to mitigate domain shift between synthetic and real-world point clouds. Finally, we utilize data mixing consistency loss to push the model to learn context-free representations. We implement and thoroughly evaluate our approach through extensive comparisons with state-of-the-art methods. Experiments on two challenging synthetic-to-real point cloud semantic segmentation tasks demonstrate that our approach achieves superior performance. Ablation studies confirm the effectiveness of the DPLF and PG-DAP modules. We release the code of our method in this paper.
\end{abstract}

\section{INTRODUCTION}

Semantic segmentation of 3D LiDAR point clouds is one of the key components of many intelligent autonomous systems such as self-driving vehicles working in dynamic, real-world environments \cite{8967762}.
To achieve scene understanding, point cloud data collected in real environments need to be densely labeled, but this process is costly and labor-intensive.
One feasible way is to leverage annotated synthetic point cloud data, as the annotations for synthetic data are automatically generated during its creation.
However, due to the domain distribution shift between synthetic environments and the real world, models trained on synthetic data often exhibit substantial performance degradation when applied to real-world point clouds.

\begin{figure}[t]
\centering
\includegraphics[width=0.9\linewidth, trim=0 40 0 30, clip]{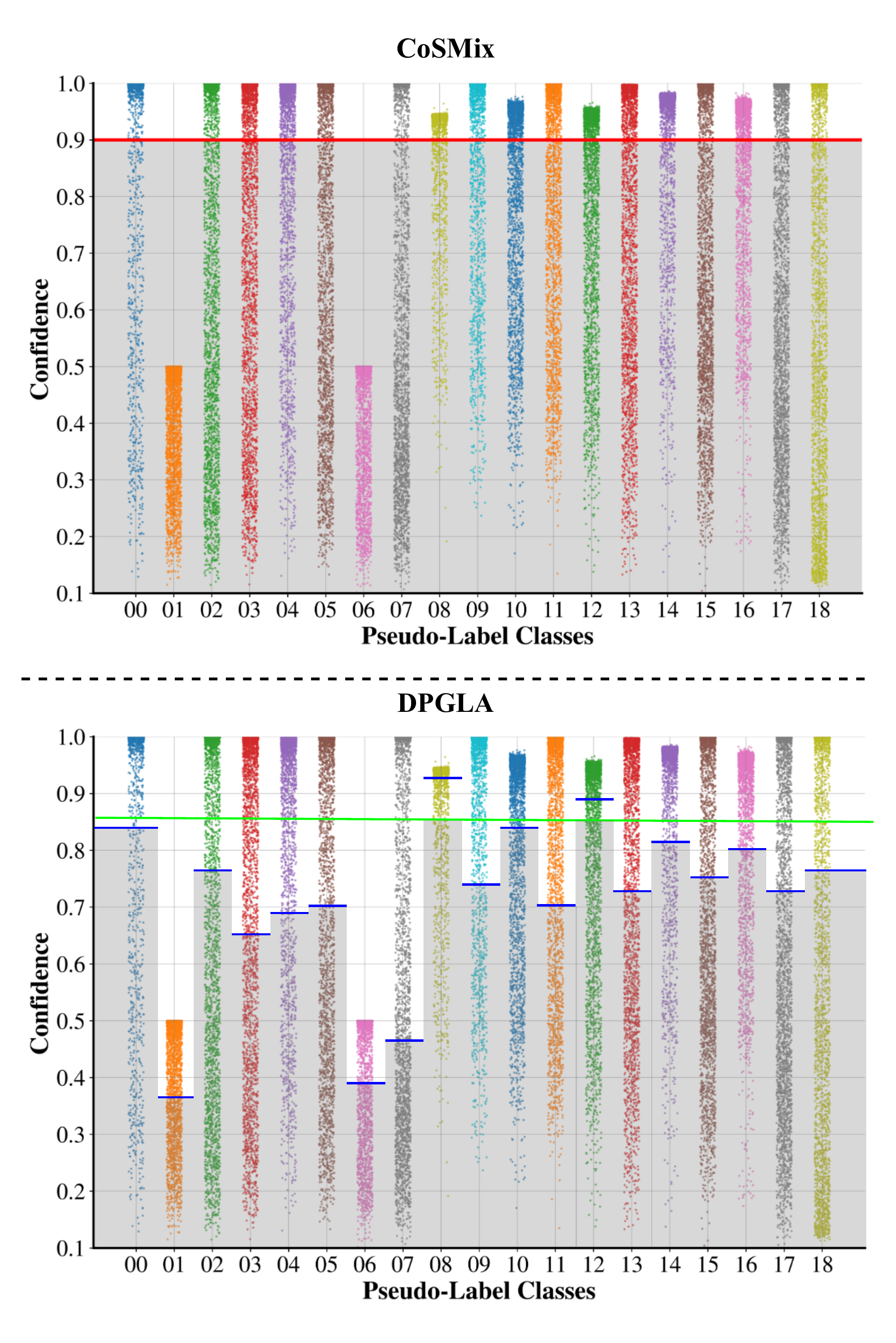}
\caption{Confidence distribution of pseudo-labels for different classes on sequence 01 of the SemanticKITTI \cite{9010727}. 
The gray area represents real point cloud data excluded from subsequent training, while the white area indicates the retained data.
The red line represents the fixed confidence threshold in CoSMix.
The green and blue lines denote global and class-specific thresholds of DPLF in DPGLA, respectively.
}
\label{fig_rdf}
\vspace{-20pt}
\end{figure}

Unsupervised Domain Adaptation (UDA) methods \cite{pmlr-v37-ganin15} are proposed to address the shift between the source and target domain, by transferring knowledge from labeled source data to unlabeled target data.
Self-training-based methods demonstrate good performance in point cloud UDA semantic segmentation by iteratively generating pseudo-labels on real point clouds and using them to retrain the model.
However, state-of-the-art methods \cite{10.1007/978-3-031-19827-4_34, 10160410, 10655228} nowadays rely on predefined or fixed confidence thresholds to filter pseudo-labels. 
Unfortunately, these simple pseudo-label filtering strategies lead to inefficient use of real data. 
As an example, we analyzed the confidence distribution of the pseudo-labels for sequence 01 in the SemanticKITTI \cite{9010727} generated by CoSMix \cite{10.1007/978-3-031-19827-4_34}, which uses a fixed confidence threshold 0.9, as shown in the upper part of \figref{fig_rdf}.
The observation shows that the fixed confidence threshold leads to severe pseudo-label filtering imbalance. Notably, almost all data with pseudo-labels of class 01 and class 06 are excluded from subsequent training, resulting in class imbalance and inefficient use of real unlabeled point clouds. Addressing this problem is one of the goals of this work.

The main contribution of this paper is a novel self-training-based method, DPGLA (Dynamic Prior-Guided LiDAR Adaptation), for accurate 3D LiDAR point cloud UDA semantic segmentation.
We achieve this by designing a Dynamic Pseudo-Label Filtering (DPLF) scheme, as illustrated in the bottom part of \figref{fig_rdf}.
DPLF employs an adaptive strategy and comprises three core components.
First, distance-based weights are assigned to the confidence of the pseudo-labels to favor denser regions, ensuring key geometric features are retained after filtering.
Then pseudo-labels are hierarchically filtered using both global and class-specific confidence thresholds. 
Finally, the global and class-specific thresholds are dynamically updated via two sets of Exponential Moving Averages (EMA), relying entirely on the statistical properties of pseudo-label confidence.

Another challenge is mitigating the input-level shift between the two domains. This shift is mainly in sparsity and noise.
Some existing methods \cite{NEURIPS2022_475b85eb} use Generative Adversarial Networks (GANs) to transform synthetic point cloud into real point cloud styles. However, these methods require significant computational resources. 
To mitigate the input-level domain shift without expensive computational overhead, we design a Prior-Guided Data Augmentation Pipeline (PG-DAP), which is based on prior knowledge and non-learned so it is more efficient. 
In PG-DAP, source and target point clouds are mixed leveraging a state-of-the-art mixing approach \cite{10205234}.
In parallel, we use a data mixing consistency loss to push the model to learn context-free representations.\\

The contributions of our paper are summarized as follows:

\begin{itemize}
    \item We propose a novel self-training-based approach for 3D LiDAR point cloud UDA semantic segmentation, achieving state-of-the-art performance on two synthetic-to-real tasks.
    \item We introduce a Dynamic Pseudo-Label Filtering (DPLF) scheme, which enables dynamically determining global and class-specific confidence thresholds for pseudo-labels while improving performance.
    \item We design a simple and efficient Prior-Guided Data Augmentation Pipeline (PG-DAP) to mitigate input-level domain shift.
    \item We release the code and implementation details at: \\ \url{https://github.com/lichonger2/DPGLA}
\end{itemize}

\section{RELATED WORK}

\begin{figure*}[t]
\includegraphics[width=\linewidth]{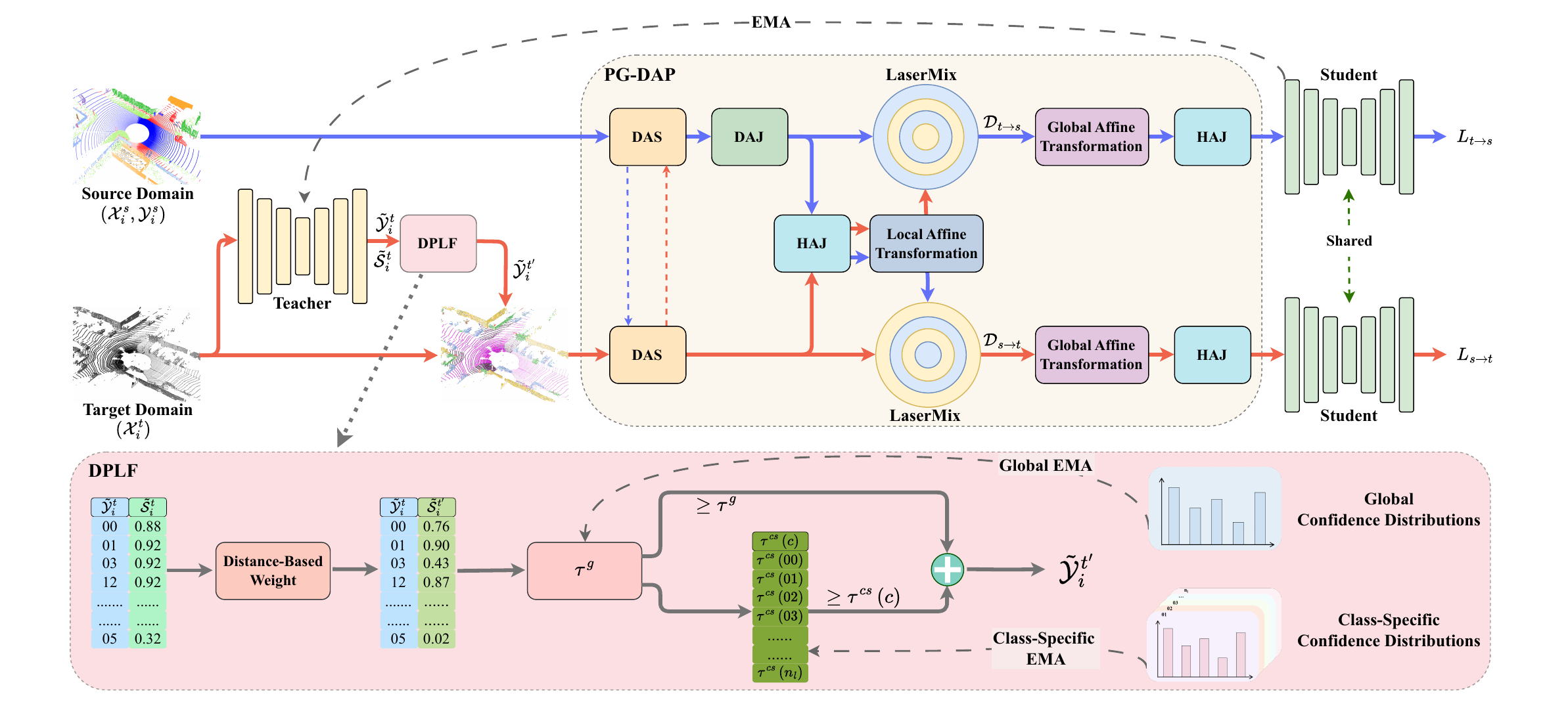}
\caption{Overview of DPGLA architecture. The model takes as input a pair of sample point clouds: one labeled from source domain $\mathcal{D}_s$ (upper branch) and one unlabeled from target domain $\mathcal{D}_t$ (bottom branch).}
\label{overflow}
\vspace{-15pt}
\end{figure*}

\noindent \textbf{Point Cloud Semantic Segmentation} assigns each point a semantic label.
Projection-based methods \cite{10.1007/978-3-031-72667-5_13, 8967762, 8793495} would first project 3D point cloud into 2D images and later utilize 2D convolutional neural networks to achieve segmentation. Following PointNet \cite{8099499}, numerous point-based methods \cite{NIPS2017_d8bf84be, 9010002, NEURIPS2022_d78ece66} have been proposed where point cloud is input directly into the network to obtain point features. 
Voxel-based methods \cite{8579059, 9495168, 10203638, 10203552} provide an alternative and popular approach. By dividing the point cloud into voxels, these methods use 3D sparse convolution to extract geometric relationships between voxels. 
MinkowskiNet \cite{8953494} is a voxel-based approach, which is the backbone network used in this study.

\noindent  \textbf{UDA} aims to train well-performing models from a labeled source domain to an unlabeled target domain by mitigating domain shift.
The mainstream methods consist of three groups: domain discrepancy alignment, adversarial learning, and self-training.
In domain discrepancy alignment, suitable metrics such as Maximum Mean Difference\cite{NEURIPS2018_ab88b157}, Central Moment Difference\cite{ZELLINGER2019174}, Maximum Density Difference \cite{9080115}, Entropy Minimization \cite{8954439} and Wasserstein Distance \cite{Liu_Han_Bai_Ge_Wang_Han_Li_You_Lu_2020} are utilized to reduce the domain discrepancy in the potential feature space.
Adversarial learning-based methods generally use Generative Adversarial Networks (GANs) \cite{10.1145/3422622} architecture. By training a domain discriminator, the features of the target domain are learned to approximate the source domain features in distribution and achieve cross-domain alignment. Adversarial learning can be performed at different stages, including feature encoding \cite{9439889, 8953759}, latent feature space \cite{pmlr-v162-rangwani22a} and output prediction \cite{8578878, 9372870}.
The self-training approach \cite{9879466,Liu_Han_Bai_Ge_Wang_Han_Li_You_Lu_2020, 9578759,9889681, 9010413} generates pseudo labels for the target domain data and involves them in later training to gradually improve the model performance. 

\noindent \textbf{Point Cloud UDA Semantic Segmentation} is valued due to it allows to use solely labeled source data when segmenting new target domain point cloud. 
Early point cloud UDA segmentation methods \cite{8793495,9341508, 9561255} project the point cloud into images, and then leverage image 2D UDA segmentation methods. 
More recent approaches can be categorized into two primary groups: adversarial learning, and self-training.

Adversarial learning methods work to reduce the domain differences between two point clouds. Complete \& Label \cite{9578920} introduces local adversarial learning to model the surface prior and uses the recovered 3D surface as the canonical domain. ePointDA \cite{ Zhao_Wang_Li_Wu_Gao_Xu_Darrell_Keutzer_2021} compensates for pixel-level domain offsets by rendering the lost noise of the synthesized LiDAR. DCF-Net \cite{9945672} proposes a category-level adversarial framework that explicitly extracts key domain private knowledge at a low-level stage. PCT \cite{Xiao_Huang_Guan_Zhan_Lu_2022} uses two generators and discriminators to transform the appearance and sparsity of the point cloud, respectively. PMAN \cite{10007866} uses prototypes (class centroids) to guide the alignment between different domains. 

Self-training methods usually employ the Mean Teacher structure \cite{NIPS2017_68053af2}. ConDA \cite{10160410} introduces regularization techniques to mitigate domain gaps, while SCT \cite{10330760} enforces consistency constraints. 
PolarMix \cite{NEURIPS2022_475b85eb}, CoSMix \cite{10.1007/978-3-031-19827-4_34} and UniMix \cite{10655228} propose various data mixing approaches to create intermediates. 
DGT-ST \cite{10658318} is a two-stage hybrid approach that uses adversarial learning to obtain pre-training weights followed by a self-training method. However, the whole process is time-consuming and requires multiple trainings. 
SALUDA \cite{10550726} considers mitigating domain shift as a surface reconstruction task.

\section{Our Approach}


\subsection{Preliminaries}\label{Preliminaries}
Following the settings of UDA, we define the source domain as the set of labeled point clouds \( \mathcal{D}_s = \{(\mathcal{X}_i^s, \mathcal{Y}_i^s)\}_{i=1}^{n_s} \), with $\mathcal{X}_i^s$ the point clouds, $\mathcal{Y}_i^s$ the labels and $n_s$ the number of samples, and the target domain \( \mathcal{D}_t = \{(\mathcal{X}_i^t)\}_{i=1}^{n_t} \) with $n_t$ unlabeled point clouds. 
Each point cloud \( \mathcal{X}_i \in \mathbb{R}^{n_i \times 3} \) consists of \( n_i \) points, where each point is represented as $\mathbf{x}_{i,j} = (x_{i,j}, y_{i,j}, z_{i,j}) \in \mathbb{R}^3$.
The corresponding point-wise semantic labels are represented as a vector \( \mathcal{Y}_i^s \in \mathbb{R}^{n_i} \).
The source and target domains follow the joint probability distributions \( p \) and \( q \). Since domain adaptation is required, we assume \( p \neq q \), indicating a distribution shift between the two domains.
Unless stated otherwise, the source point cloud refers to synthetic data, and the target point cloud to real data.
Our goal is to train a network capable of predicting accurate semantic labels \( \hat{\mathcal{Y}}^t \) for new samples from \( \mathcal{D}_t \). 

\subsection{Architecture}\label{Architecture}
The architecture of DPGLA follows the Mean Teacher model \cite{NIPS2017_68053af2}, as illustrated in \figref{overflow}. 
The student and teacher networks are represented as $\Phi_{\theta}$ and $\Phi_{\theta'}$ with learnable parameters $\theta$ and $\theta'$, respectively.

During training, the teacher network $\Phi_{\theta'}$ generates pseudo-labels for the unlabeled target samples. The teacher network is initialized by supervised pre-training using only the labeled source domain:
\begin{equation}
\mathcal {L}_{pretrain} = \mathcal {L}_{SD}(\Phi _{\theta'}(\mathcal{X}_i^s), \mathcal{Y}_i^s),
\label{eq:source_loss}
\end{equation}
where the Soft Dice loss \cite{7785132} is employed as \(\mathcal {L}_{SD}\).
Pseudo-labels are filtered through Dynamic Pseudo-Label Filtering (DPLF, \secref{DPLF}), retaining high-quality ones for subsequent training.
We leverage a Prior-Guided Data Augmentation Pipeline (PG-DAP, \secref{sec:data_aug}) to mitigate the shift of the two domains at the input level. 
As part of PG-DAP, we adopt the LaserMix framework \cite{10205234} for data mixing, where source and target point clouds are partitioned into regions, and selected regions are exchanged between the two domains. 
Let \( \mathcal{D}_{t \to s} \) be the mixed point cloud obtained from the upper branch, and \( \mathcal{D}_{s \to t} \) be the mixed point cloud obtained from the bottom branch.
Finally, the student network $\Phi_{\theta}$ updates $\theta$ from the processed point clouds by optimizing the overall objective loss function, while the teacher network parameters $\theta'$ is updated by using EMA(\secref{DMC} and \secref{update}).

In the inference phase, the student network $\Phi_{\theta}$ assigns a predicted semantic label to each point of the new data in the target domain.

\subsection{Dynamic Pseudo-Label Filtering}\label{DPLF}
We argue that pseudo-label filtering plays a crucial role in the self-training-based point cloud UDA segmentation method, as pseudo-labels provide potentially valuable supervised information.
However, using fixed confidence thresholds for filtering can result in inefficient utilization of target domain data and exacerbate class imbalance, ultimately degrading performance.
Therefore, we propose a Dynamic Pseudo-Label Filtering scheme (DPLF, pale red box in \figref{overflow}) that adaptively adjusts the confidence threshold for each class during the training process.
Original pseudo-labels \(\tilde{\mathcal{Y}}_i^t\) and confidence scores \(\tilde{\mathcal{S}}_i^t\) are produced by the pretrained teacher network:
\begin{equation}
\tilde{\mathcal{Y}}_i^t= \arg\max(\mathrm{softmax}(\Phi _{\theta '}(\mathcal{X}_i^t))),
 \end{equation}
 \begin{equation}
\tilde{\mathcal{S}}_i^t= \max(\mathrm{softmax}(\Phi _{\theta '}(\mathcal{X}_i^t))).
 \end{equation}
Next, we introduce three sub-processes of DPLF, which sequentially filter out pseudo-labels for subsequent training:

\noindent \textbf{Distance-Based Weight.}\enspace
As the distance increases, LiDAR point clouds become increasingly sparse, all pseudo-labels are not treated equally.
We assign distance-based weights \({w}_i^d \) to pseudo-labels, prioritizing points in denser regions closer to the LiDAR. 
This ensures that denser instances have higher retention probability, preserving geometric features for training after the filtering process as:
\begin{equation}
{w}_{i,j}^d = {\exp\left(-\alpha \cdot \tilde{d}_{i,j}^t \right)},
\end{equation}
\begin{equation}
\tilde{\mathcal{S}}_{i,j}^{t\prime} = \tilde{\mathcal{S}}_{i,j}^t \cdot w_{i,j}^d,
\end{equation}
where \(\alpha\) is a regulatory factor, \(\tilde{d}_{i,j}^t\) is the normalized Euclidean distance between the point $\mathbf{x}_{i,j}^t$ and the LiDAR in target domain and $\tilde{\mathcal{S}}_{i,j}^t $ is the original confidence score of pseudo-label a point.

\noindent \textbf{Pseudo-Label Filtering.}\enspace
To ensure that the pseudo-labels used for model training are reliable and class-balanced, we apply a hierarchical filtering mechanism based on the global threshold and class-specific thresholds.
First, to prevent distant noise points or highly unreliable pseudo-labels from distorting the overall distribution properties, the bottom 1\% of pseudo-labels with the lowest confidence scores are rejected by default, i.e. marked as \textit{unknown}s ($-1$ label). All points labeled with $-1$ label are excluded from subsequent training. Then pseudo-labels with confidence scores above the global threshold \(\tau^g\) are retained. The remainder are compared to their respective class-specific thresholds $\tau^{cs}\left(c\right)$, where \(c\) denotes a specific class.
Practically, the smaller threshold decides whether a pseudo-label is retained, i.e.:

\begin{equation}
\tilde{\mathcal{Y}}_i^{t\prime} =
\begin{cases}
\tilde{\mathcal{Y}}_{i,j}^t, & \tilde{\mathcal{S}}_{i,j}^{t\prime}\left(c\right) \geq \min (\tau^g, \tau^{cs}\left(c\right)), \quad c \in C_{l}, \\[6pt]
-1, & \text{otherwise},
\end{cases}
\end{equation}
where $\tilde{\mathcal{Y}}_{i,j}^t$ is the point's original pseudo-label and \(C_{l}\) represents the set of all \(n_{l}\) classes in target domain.
The goal here is to refine the original set of pseudo-labels, $\tilde{\mathcal{Y}}_i^t$, producing a new filtered set, $\tilde{\mathcal{Y}}_i^{t\prime}$. Pseudo-labels not meeting the filtering criteria are assigned with the $-1$ label.

We define the global threshold and class-specific thresholds based on two statistical properties of respective confidence distributions: the mean (\(\mu^g\) / \(\mu^{cs}\)) and variance (\(\sigma^g\) / \(\sigma^{cs}\)):
\begin{equation}
\tau^g = \mu^g + \sigma^g,
\end{equation}
\begin{equation}
\tau^{cs}\left(c\right) = \mu^{cs}\left(c\right) - \sigma^{cs}\left(c\right).
\end{equation}
These thresholds are updated dynamically, as described below.
It is worth noting that we used different operations for these two statistical properties when designing the global and class-specific thresholds for special considerations. 

In self-training methods, ``overconfident'' \cite{NEURIPS2021_c1fea270} problem arises as training progresses, where incorrect pseudo-labels tend to be assigned high confidence, leading to an increasing skew in the overall confidence distribution. 
To tackle this, we ensure that the global threshold becomes progressively stricter over time by summing the mean \(\mu^g\) and variance \(\sigma^g\). 
This helps mitigate the negative impact of ``overconfident'' problem. 
The variance serves as a hedging factor as it captures changes in skewness. 

For class-specific thresholds, different classes exhibit varying uncertainty levels, with minority classes generally having higher uncertainty. By subtracting variance \(\sigma^{cs}\left(c\right)\) from the mean \(\mu^{cs}\left(c\right)\), we provide more relaxed thresholds for highly uncertain classes, promoting class balance. The variance acts as a class-balancing factor, preventing minority classes from being excessively filtered.

\noindent \textbf{Dynamic Threshold Update.}  \enspace
To respond to changes in the confidence distribution, the mean and variance must be dynamically updated to adjust the thresholds. 
However, due to the large-scale data, computing the confidence distribution of all unlabeled data at each time step would be highly time-consuming. 
Instead, we employ two sets of Exponential Moving Averages (EMA), including global EMA and class-specific EMA, to estimate the statistical properties at \(t\)-th time iteration, by using only the point clouds that belong to the current iteration batch. 
The update process performed every $\gamma$ iterations, denoted as:
\begin{equation}
\begin{split}
\mu^g &= \lambda^g \cdot \mu_t^g +(1 - \lambda^g) \mu_{t-1}^g, \\
\sigma^g &= \lambda^g \cdot \sigma_t^g +(1 - \lambda^g) \sigma_{t-1}^g,
\end{split}
\end{equation}
\begin{equation}
\begin{split}
\mu^{cs}\left(c\right) &= \lambda^{cs} \cdot \mu_t^{cs}\left(c\right) + (1 - \lambda^{cs}) \mu_{t-1}^{cs}\left(c\right), \\
\sigma^{cs}\left(c\right) &= \lambda^{cs} \cdot \sigma_t^{cs}\left(c\right) + (1 - \lambda^{cs}) \sigma_{t-1}^{cs}\left(c\right),
\end{split}
\end{equation}
where \(\lambda^{g}, \lambda^{cs}\in [0,1)\) are the momentum coefficient of the global EMA and the class-specific EMA respectively.

\subsection{Prior-Guided Data Augmentation Pipeline}\label{sec:data_aug}
The main input-level domain shift between synthetic and real point clouds comes from varying degrees of sparsity and noise.
If left unaddressed, this shift can lead to negative transfer, ultimately degrading UDA segmentation performance on the target domain.
Adversarial learning-based methods usually train a Generative Adversarial Networks (GANs) to align the style of the source point cloud with the target domain. 
However, these methods require additional computational resources and often demand multiple training cycles due to the instability of GANs.
To mitigate the input-level domain shift without expensive computational
overhead, we propose a simple and efficient prior-guided data augmentation pipeline (PG-DAP) by leveraging prior knowledge, requiring no additional learning.
In our pipeline, we employ spatially prior-guided LaserMix \cite{10205234} and adopt the widely used strategy of combining local and global affine transformations, as demonstrated in previous works such as CoSMix \cite{10.1007/978-3-031-19827-4_34}.
More importantly, three novel techniques are introduced: Density-Aware Sampling (DAS), Distance-Aware Jitter (DAJ), and Height-Aware Jitter (HAJ).

\noindent \textbf{Density-Aware Sampling (DAS).} \enspace
The distribution of points at different distances is irregular in the two domains \cite{10658318}.  
To address this, DAS employs a soft-sampling mechanism that adjusts the number of points in each distance interval, aligning the density distribution between the two domains.
First, the point clouds are partitioned into \( n \) and \( m \) bins in the source and target domains, respectively, using a step size of \( \Delta d \):
\begin{equation}
\begin{split}
b_u^s &= \{(u-1) \Delta d \leq \|\mathbf{x}_{i,j}^s\|_2 < u \Delta d \}, \\
b_v^t &= \{(v-1) \Delta d \leq \|\mathbf{x}_{i,j}^t\|_2 < v \Delta d \}.
\end{split}
\end{equation}
Here, \( b_u^s \) and \( b_v^t \) denote the sets of bins in the source and target domains, respectively, where \( u \in \{1, 2, \dots, n\} \) and \( v \in \{1, 2, \dots, m\} \) represent the bin indices.
We limit our adjustment to the corresponding bin pairs in both domains, specifically the first \( k \) bins, where \( k \in \{1, 2, \dots, \min(n, m)\} \).
The number of points in $b_k^s$ and $b_k^t$ is denoted as \( N_k^s \) and \( N_k^t \), respectively.  
To align the distributions, we determine the number of sampled points \( \hat{N}_k \) in each bin as:
\begin{equation}
\hat{N}_k = \min(N_k^s, N_k^t) \cdot \xi,
\end{equation}
where \( \xi \) is a soft factor randomly sampled from the range $[1 - \epsilon, 1 + \epsilon]$ and \( \epsilon > 0 \).  
For each interval set $b_k^s$ and $b_k^t$, we randomly subsample the sparser set, retaining $\hat{N}_k$ points, to prevent overrepresentation of either domain.

\noindent \textbf{Distance-Aware Jitter (DAJ).}\enspace
In real-world LiDAR data acquisition, geometric noise increases with distance\cite{10839256} due to signal attenuation and environmental interference.
Based on this prior knowledge, DAJ adapts the clean source domain by introducing noise that mimics real-world data. This is achieved by perturbing each source domain point $\mathbf{x}_{i,j}$ as follow:
\begin{equation}
{\mathbf{x}_{i,j}}^{\prime} = \mathbf{x}_{i,j} + \mathbf{n}_{i,j}^{DAJ},
\end{equation}
where $\mathbf{n}_{i,j}^{DAJ}$ is a zero-mean Gaussian random noise sampled from:
\begin{equation*}
\mathcal{N} \left( \mathbf{0}, \left( \sigma_{\text{min}} + (\sigma_{\text{max}} - \sigma_{\text{min}}) \cdot \sqrt{\tilde{d}_{i,j}} \cdot \xi \right)^2 \mathbf{I} \right),
\end{equation*}
where \(\tilde{d}_i^s\) is the normalized Euclidean distance between the point and the LiDAR in source domain, \(\sigma_{\text{min}}\) and \(\sigma_{\text{max}}\) are parameters that regulate the maximum and minimum noise, respectively
and $\mathbf{I}$ is the $3 \times 3$ identity matrix.

\noindent \textbf{Height-Aware Jitter (HAJ).}\enspace
A heuristic prior from \cite{10205234} suggests that LiDAR beams tend to capture more instances of certain classes in specific pitch angle ranges, as some instances cluster more densely in those ranges.
A straightforward example is that \textit{road} points are often captured by LiDAR beams with smaller pitch angles, while \textit{car} points are mostly detected by beams with moderate pitch angles.
We extend this insight to the height dimension.
Observing point clouds in both domains, instances at different heights exhibit distinct noise characteristics due to LiDAR sampling and object geometry. 
Low-height instances (e.g., \textit{roads}, \textit{terrain}) mainly show noise discrepancies in the XY plane, whereas high-height instances (e.g., \textit{buildings}) experience noise differences primarily along the Z-axis.
Based on this observation, we propose HAJ, which mitigates the noise difference between two domains by applying structured noise at different height ranges.
The HAJ points perturbation is defined as follows: 
\begin{equation}
{\mathbf{x}_{i,j}}' = \mathbf{x}_{i,j} + \mathbf{n}_{i,j}^{HAJ} ,
\end{equation}
with:
\begin{equation}
\quad \mathbf{n}_{i,j}^{HAJ} \sim \mathcal{N}(\mathbf{0}, \sigma_{\text{HAJ}} \cdot \mathbf{W}_{i,j} \cdot \xi)^2 \mathbf{I}),
\end{equation}
\begin{equation}
\mathbf{W}_{i,j} = \text{diag}(w_{i,j}^x, w_{i,j}^y, w_{i,j}^z),
\end{equation}
where \(\mathbf{W}_{i,j}\) is a noise modulation matrix, composed of weights (\( w_{i,j}^x, w_{i,j}^y, w_{i,j}^z \)) that control the noise intensity in each of the three spatial directions. 
We define these three weights based on the normalized height \( \tilde{z}_{i,j} \), using two height thresholds, \(h_{\text{low}}\) and \(h_{\text{high}}\), as follows:
\begin{equation}
\begin{split}
w_{i,j}^x &= w_{i,j}^y = \mathds{1}( \tilde{z}_{i,j} < h_{\text{high}}),\\
w_{i,j}^z &= \mathds{1}(\tilde{z}_{i,j} > h_{\text{low}}).\\ 
\end{split}
\end{equation}
where \(\mathds1(\cdot)\) is an indicator function, which returns 1 if the given condition is met and 0 otherwise.

\subsection{Data Mixing Consistency}\label{DMC}
In point cloud UDA segmentation, the network learns contextual dependencies, which can lead to negative transfer in the target domain.
Therefore, we leverage a data mixing consistency loss to guide the network in learning context-free feature representations.

LaserMix \cite{10205234} is a recent point cloud data mixing approach that divides two point clouds into different regions based on pitch angles before mixing. In our method, we directly adopt this approach.
The mixed domain \(\mathcal{D}_m\) has $n_m$ point clouds, which is denoted as:
\begin{equation}
\begin{aligned}
\mathcal{D}_m &= \{(\mathcal{X}_i^m, \mathcal{Y}_i^m)\}_{i=1}^{n_m}, \\
\mathcal{X}_i^m &= \mathcal{X}_i^s[r_m] \cup \mathcal{X}_i^t[r_m], \\
\mathcal{Y}_i^m &= \mathcal{Y}_i^s[r_m] \cup \tilde{\mathcal{Y}}_i^{t\prime}[r_m],
\end{aligned}
\end{equation}
 where \( m \in \{ t \to s, s \to t \} \) indicates the mixing direction, with \( t \to s \) denoting mixed point cloud from the upper branch and \( s \to t \) denoting mixed point cloud from the bottom branch.
The subsets \( \mathcal{X}_i^s[r_m] \) and \( \mathcal{X}_i^t[r_m] \) are sampled point clouds from the source and target domain, respectively, with \( \mathcal{Y}_i^s[r_m] \) as ground-truth labels and \(\tilde{\mathcal{Y}}_i^{t\prime}[r_m] \) as pseudo-labels.
To ensure that the model learns robust context-free feature representations from mixed data, we introduce a data mixing consistency loss:
\begin{equation}
\mathcal{L}_{dmc} = \mathcal {L}_{CE} \left(\mathcal{Y}_i^m, \Phi _{\theta}(\mathcal{X}_i^m)\right),
\end{equation}
where $\mathcal {L}_{CE}$ is Cross Entropy loss.
This loss pushes the consistency of the student network's predictions for mixed samples with their corresponding labels or pseudo-labels, unperturbed by contextual changes. 
Encouraging the model to learn more context-free representations improves its ability to adapt across domains.
\subsection{Network Update}\label{update}
We employ the Mean Teacher architecture (\secref{Architecture}) to enhance knowledge transfer acquired throughout the training process in mixed domains. 
We define the segmentation loss $\mathcal {L}_{seg}$ for two branches as:
\begin{equation}
\mathcal {L}_{seg} = \mathcal {L}_{SD} \left(\mathcal{Y}_i^m, \Phi _{\theta}(\mathcal{X}_i^m)\right),
\end{equation}
where Soft Dice segmentation loss \cite{7785132} is implemented as \(\mathcal {L}_{SD}\).
The parameters $\theta$ of student network $\Phi_{\theta}$ are updated based on the following overall objective loss function:
\begin{equation}
\mathcal{L} = w_{seg}\mathcal {L}_{seg} + w_{dmc} \mathcal {L}_{dmc},
\end{equation}
where $w_{seg}$ and $w_{dmc}$ represent the loss weight for $\mathcal {L}_{seg}$ and $\mathcal {L}_{dmc}$, respectively.
On the other hand, we update the parameters $\theta'$ of teacher network $\Phi_{\theta'}$ every $\beta$ iterations at \(t\)-th time step by applying the Exponential Moving Average (EMA):
\begin{equation}
\theta'_t = \lambda^N \cdot \theta'_{t-1}  +  (1 - \lambda^N) \theta_t,
\end{equation}
where $\lambda^N \in [0,1)$ is a momentum coefficient.

\begin{table*}[ht]
\centering
\caption{Comparison among the state-of-the-art point cloud UDA semantic segmentation methods on the validation set (sequence 08) of SemanticKITTI.}
\label{semantickitti}
\renewcommand{\arraystretch}{0.9}
\definecolor{lightyellow}{RGB}{255, 255, 200}
\definecolor{lightblue}{RGB}{0, 0, 0}
\definecolor{golden}{RGB}{0,0, 0}
\huge
\resizebox{\textwidth}{!}{
\begin{tabular}{r|ccccccccccccccccccc|c}
\toprule
Mehod (year) & \rotatebox{90}{car}  & \rotatebox{90}{bi.cle} & \rotatebox{90}{mt.cle} & \rotatebox{90}{truck} & \rotatebox{90}{oth-v.} & \rotatebox{90}{pers.} & \rotatebox{90}{b.clst} & \rotatebox{90}{m.clst} & \rotatebox{90}{road} & \rotatebox{90}{park.} & \rotatebox{90}{sidew.} & \rotatebox{90}{oth-g.} & \rotatebox{90}{build.} & \rotatebox{90}{fence} & \rotatebox{90}{veget.} & \rotatebox{90}{trunk} & \rotatebox{90}{terra.} & \rotatebox{90}{pole} & \rotatebox{90}{traff.} & mIoU \\ 
\midrule
\rowcolor{lightyellow} Source      & 42.0 & 5.0    & 4.8    & 0.4   & 2.5    & 12.4  & 43.3   & 1.8    & 48.7 & 4.5   & 31.0   & 0.0    & 18.6   & 11.5  & 60.2   & 30.0  & 48.3  & 19.3 & 3.0    & 20.4 \\ 
\midrule
ADDA \cite{8099799} ['17]  & 52.5 & 4.5    & 11.9   & 0.3   & 3.9    & 9.4   & 27.9   & 0.5    & 52.8 & 4.9   & 27.4   & 0.0    & 61.0   & 17.0  & 57.4   & 34.5  & 42.9  & 23.2 & 4.5    & 23.0 \\ 
Ent-Min \cite{8954439} ['19] & 58.3 & 5.1   & 14.3   & 0.3   & 1.8    & 14.3  & 44.5   & 0.5    & 50.4 & 4.3   & 34.8   & 0.0    & 48.3   & \underline{19.7}  & 67.5   & 34.8  & \underline{52.0}  & 33.0 & 6.1    & 25.8 \\ 
ST \cite{9010413} ['19]  & 62.0 & 5.0    & 12.4   & 1.3   & 9.2    & 16.7  & 44.2   & 0.4    & 53.0 & 2.5   & 28.4   & 0.0    & 57.1   & 18.7  & 69.8   & 35.0  & 48.7  & 32.5 & 6.9    & 26.5 \\ 
PCT \cite{10330760} ['22]   & 53.4 & 5.4    & 7.4    & 0.8   & \underline{10.9}   & 12.0  & 43.2   & 0.3    & 50.8 & 3.7   & 29.4   & 0.0    & 48.0   & 10.4  & 68.3   & 33.1  & 40.0  & 29.5 & 6.9    & 23.9 \\ 
ST-PCT \cite{10330760} ['22] & 70.8 & 7.3    & 13.1   & 1.9   & 8.4    & 16.2  & 44.0   & 0.6    & 56.4 & 4.5   & 31.8   & 0.0    & \underline{66.7}   & \textbf{23.7}  & \textbf{73.3}   & 34.6  & 48.4  & 39.4 & 11.7   & 28.9 \\ 
PolarMix \cite{NEURIPS2022_475b85eb} ['22] & \underline{76.3} & \textbf{8.4} & 17.8 & 3.9 & 6.0 & \underline{26.6} & 40.8 & \underline{15.9} & 70.3 & 0.0 & 44.4 & 0.0 & \textbf{68.4} & 14.7 & 69.6 & \underline{38.1} & 37.1 & 40.6 & 10.6 & 31.0 \\ 
CoSMix \cite{10.1007/978-3-031-19827-4_34}  ['22] & 75.1 & 6.8 & \underline{29.4} & \underline{27.1} & \textbf{11.1} & 22.1 & 25.0 & \textbf{24.7} & \underline{79.3} & \textbf{14.9} & \underline{46.7} & \underline{0.1} & 53.4 & 13.0 & 67.7 & 31.4 & 32.1 & 37.9 & 13.4 & \underline{32.2} \\ 
SALUDA \cite{10550726} ['24] & 67.0 & \underline{7.7} & 14.4 & 1.3 & 5.2 & 24.1 & \underline{52.6} & 2.7 & 52.5 & 10.5 & 44.1 & \textbf{0.4} & 51.8 & 13.6 & 69.7 & \textbf{40.5} & \textbf{56.5} & \textbf{45.0} & \underline{14.3} & 30.2 \\ 
\midrule
\textbf{DPGLA (Ours)} & \textbf{79.3} & 6.1 & \textbf{42.1} & \textbf{43.2} & 10.3 & \textbf{34.3} & \textbf{53.7} & 5.9 & \textbf{81.8} & \underline{12.0} & \textbf{51.6} & 0.0 & 57.8 & 13.2 & \underline{71.7} & 35.6 & 37.9 & \underline{44.6} & \textbf{24.2} & \textbf{37.1} \\ 
\textit{Improv.} ↑ 
& \textit{\textcolor{lightblue}{+37.3}} & \textit{\textcolor{lightblue}{+1.1}} & \textit{\textcolor{lightblue}{+37.3}} 
& \textit{\textcolor{lightblue}{+42.8}} & \textit{\textcolor{lightblue}{+7.8}} & \textit{\textcolor{lightblue}{+21.9}} 
& \textit{\textcolor{lightblue}{+10.4}} & \textit{\textcolor{lightblue}{+4.1}} & \textit{\textcolor{lightblue}{+33.1}} 
& \textit{\textcolor{lightblue}{+7.5}} & \textit{\textcolor{lightblue}{+20.6}} & \textit{\textcolor{lightblue}{+0.0}} 
& \textit{\textcolor{lightblue}{+39.2}} & \textit{\textcolor{lightblue}{+1.6}} & \textit{\textcolor{lightblue}{+11.5}} 
& \textit{\textcolor{lightblue}{+5.6}} & \textit{\textcolor{golden}{-10.4}} & \textit{\textcolor{lightblue}{+25.3}} 
& \textit{\textcolor{lightblue}{+21.2}} & \textit{\textcolor{lightblue}{+16.7}} \\ 

\bottomrule
\end{tabular}
}
\end{table*}

\begin{table*}[ht]
\centering
\caption{Comparison among the state-of-the-art point cloud UDA semantic segmentation methods on the validation set (sequence 03) of SemanticPOSS}
\label{semanticposs}
\renewcommand{\arraystretch}{0.9}
\definecolor{lightyellow}{RGB}{255, 255, 200}
\definecolor{lightblue}{RGB}{0, 0, 0}
\definecolor{golden}{RGB}{0 , 0, 0}
\resizebox{\textwidth}{!}{
\begin{tabular}{r|ccccccccccccc|c}
\toprule
Method (year) & pers. & rider & car & trunk & plants & traf. & pole & garb. & buil. & cone. & fence & bike & grou. & mIoU \\
\midrule
\rowcolor{lightyellow} Source & 3.7 & 25.1 & 12.0 & 10.8 & 53.4 & 0.0 & 19.4 & 12.9 & 49.1 & 3.1 & 20.3 & 0.0 & 59.6 & 20.7 \\
\midrule
ADDA \cite{8099799} ['17] & 27.5 & 35.1 & 18.8 & 12.4 & 53.4 & 2.8 & 27.0 & 12.2 & 64.7 & 1.3 & 6.3 & 6.8 & 55.3 & 24.9 \\
Ent-Min \cite{8954439} ['19] & 24.2 & 32.2 & 21.4 & 18.9 & 61.0 & 2.5 & 36.3 & 8.3 & 56.7 & 3.1 & 5.3 & 4.8 & 57.1 & 25.5 \\
ST \cite{9010413} ['19] & 23.5 & 31.8 & 22.0 & 18.9 & 63.2 & 1.9 & \textbf{41.6} & 13.5 & 58.2 & 1.0 & 9.1 & 6.8 & 60.3 & 27.1 \\
PCT \cite{10330760} ['22] & 13.0 & 35.4 & 13.7 & 10.2 & 53.1 & 1.4 & 23.8 & 12.7 & 52.9 & 0.8 & 13.7 & 1.1 & 66.2 & 22.9 \\
ST-PCT \cite{10330760} ['22] & 28.9 & 34.8 & 27.8 & 18.6 & 63.7 & 4.9 & \underline{41.0} & 16.6 & 64.1 & 1.6 & 12.1 & 6.6 & 63.9 & 29.6 \\
PolarMix \cite{NEURIPS2022_475b85eb} ['22] & 32.6 & 39.1 & 25.0 & 11.9 & 64.2 & 5.8 & 29.6 & 15.3 & 44.8 & 13.3 & 23.8 & \underline{10.7} & \textbf{79.0} & 30.4 \\ 
CoSMix \cite{10.1007/978-3-031-19827-4_34} ['22] & 55.8 & \underline{51.4} & 36.2 & 23.5 & \underline{71.3} & \underline{22.5} & 34.2 & 28.9 & \underline{66.2} & 20.4 & \underline{24.9} & 10.6 & \underline{78.7} & 40.4 \\
SALUDA \cite{10550726} ['24] & \textbf{59.9} & \textbf{54.6} & \textbf{59.2} & \textbf{33.7} & 69.8 & 14.9 & 40.9 & \underline{30.8} & 64.5 & \textbf{26.2} & 22.1 & 2.7 & 78.0 & \underline{42.9} \\
\midrule
\textbf{DPGLA (Ours)} & \underline{57.4} & 45.1 & \underline{36.3} & \underline{26.2} & \textbf{77.7} & \textbf{38.0} & 21.1 & \textbf{36.7} & \textbf{73.5} & \underline{25.1} & \textbf{50.1} & \textbf{40.0} & 75.5 & \textbf{46.4} \\
\textit{Improv.} ↑ 
& \textit{\textcolor{lightblue}{+53.7}} & \textit{\textcolor{lightblue}{+20.0}} & \textit{\textcolor{lightblue}{+24.3}} 
& \textit{\textcolor{lightblue}{+15.4}} & \textit{\textcolor{lightblue}{+24.3}} & \textit{\textcolor{lightblue}{+38.0}} 
& \textit{\textcolor{lightblue}{+1.7}} & \textit{\textcolor{lightblue}{+23.8}} & \textit{\textcolor{lightblue}{+24.4}} 
& \textit{\textcolor{lightblue}{+22.0}} & \textit{\textcolor{lightblue}{+29.8}} & \textit{\textcolor{lightblue}{+40.0}} 
& \textit{\textcolor{lightblue}{+15.9}} & \textit{\textcolor{lightblue}{+25.7}} \\
\bottomrule
\end{tabular}
}
\vspace{-10pt}
\end{table*}

\begin{table}[ht]
\centering
\caption{Ablation studies on SemanticPOSS}
\label{ablation}
\renewcommand{\arraystretch}{0.9}
\definecolor{grayrow}{RGB}{255, 255, 200} 
\definecolor{lightblue}{RGB}{155, 233, 233}
\small
\resizebox{\linewidth}{!}{  
\begin{tabular}{c|cccc|c|c|c}
    \toprule
    \textbf{} & \multicolumn{4}{c|}{\textbf{PG-DAP}} & \textbf{DPLF} & $\boldsymbol{\mathcal{L}}_{\boldsymbol{dmc}}$ & \textbf{mIoU} \\
    \textbf{} & Lasermix & DAS & DAJ & HAJ &  &  &  \\
    \midrule
    \rowcolor{grayrow} Source & - & - & - & - & - & - & 20.7 \\
    \midrule
    \rowcolor{lightblue} Baseline & \checkmark &  &  &  &  &  & 36.1 \\
    \midrule
    A & \checkmark & \checkmark &  &  &  &  & 36.5 \\
    B & \checkmark &  & \checkmark &  &  &  & 37.7 \\
    C & \checkmark &  &  & \checkmark &  &  & 37.8 \\
    D & \checkmark & \checkmark & \checkmark &  &  &  & 38.1 \\
    E & \checkmark & \checkmark & \checkmark & \checkmark &  &  & 39.9 \\
    F & \checkmark & \checkmark & \checkmark & \checkmark &  & \checkmark & 40.6 \\
    G & \checkmark & \checkmark & \checkmark & \checkmark & \checkmark &  & 44.6  \\
    \midrule
    \textbf{Full} & \checkmark & \checkmark & \checkmark & \checkmark & \checkmark & \checkmark & \textbf{46.4} \\
    \bottomrule
\end{tabular}
}
\vspace{-15pt}
\end{table}

\section{EXPERIMENTAL EVALUATION} 
The experimental evaluation aims to demonstrate that our approach (i) outperforms the state-of-the-art in the task of point cloud UDA semantic segmentation, and (ii) improves the performance through each proposed module.
\subsection{Datasets and Evaluation Metrics}
\noindent \textbf{SynLiDAR} (\textbf{SL}) \cite{Xiao_Huang_Guan_Zhan_Lu_2022} is a synthetic point cloud dataset, consisting of 198,396 LiDAR scans with 32 semantic classes annotated. The dataset is simulated using a 64-beam LiDAR. Following the official split \cite{Xiao_Huang_Guan_Zhan_Lu_2022}, 19,840 scans are used for training and 1,976 for validation. In all our experiments, we used SL as the source domain dataset.

\noindent \textbf{SemanticKITTI} (\textbf{SK}) \cite{9010727} is a real-world LiDAR segmentation dataset. It was captured using a 64-beam LiDAR, comprising 43,552 LiDAR scans with 19 semantic classes annotated. Following the official protocol \cite{9010727}, sequence 08 is used for validation, while the remaining sequences (00-10, excluding 08) are used for training.

\noindent \textbf{SemanticPOSS} (\textbf{SP}) \cite{9304596} is a real-world point cloud dataset collected using a 40-beam LiDAR, containing 2,988 scans with 14 semantic classes annotated. This dataset is different in spatial distribution from the SemanticKITTI. Following the official benchmark \cite{9304596}, sequence 03 is used for validation, with the remaining sequences used for training.\\

\noindent \textbf{Evaluation Metrics.}\enspace 
Intersection over Union (IoU) \cite{10.1007/978-3-319-50835-1_22} is used as the evaluation metric to validate our method. 
Following the typical evaluation protocol \cite{Xiao_Huang_Guan_Zhan_Lu_2022}, we compute and report the IoU score for each class. The mean Intersection over Union (mIoU) is then computed by averaging the IoU scores across all classes. IoU and mIoU scores are presented as percentage (\%).

\subsection{Implementation Details}
We implemented our method in PyTorch and conducted experiments on an NVIDIA A40 GPU.
For a fair comparison, we adopt MinkowskiNet \cite{8953494} as both teacher and student networks, aligning with the architecture used in other state-of-the-art approaches\cite{10330760, 10550726, NEURIPS2022_475b85eb, 10.1007/978-3-031-19827-4_34}.\\

\noindent \textbf{PG-DAP Parameters.}\enspace  
All experiments use the same set of method parameters in PG-DAP.  
The soft factor is defined as \( \xi \in [0.9, 1.1] \) with \( \epsilon = 0.1 \).  
For local affine transformations, rigid rotation around the z-axis and uniform scaling along all axes are applied, with rotation constrained to \([- \frac{\pi}{2}, \frac{\pi}{2}]\) and scaling within \([0.95, 1.05]\).  
For global affine transformations, rigid rotation, translation, and scaling are performed along all three axes.  
In LaserMix, the number of regions is randomly chosen from a list $[3, 4, 5, 6]$.  
For DAS, the separation distance is set to \( \Delta d = 5 \).  
For DAJ, the parameters are defined as \( \sigma_{\text{min}}=0.005 \) and \( \sigma_{\text{max}}=0.05 \), with range constrained within \([-0.1, 0.1]\)\,m.  
For HAJ, \( \sigma_{\text{HAJ}}=0.002 \), \( h_{\text{low}}=0.2 \), \( h_{\text{high}}=0.8 \) are used, with range limited to \([-0.1, 0.1]\)\,m.

\noindent \textbf{DPLF Parameters.}\enspace  
In all experiments, the parameter \( \alpha \) is set to 0.5.  
In \textbf{SL} \(\to\) \textbf{SK} adaptation,  
when \( t \leq 500 \), the parameters are initialized as \( \lambda^g = \lambda^{cs} = \frac{1}{t + 1} \) and \( \gamma = 1 \), allowing the model to quickly capture the initial confidence distribution.  
When \( t > 500 \), the values are adjusted to \( \lambda^g = 0.1 \), \( \lambda^{cs} = 0.01 \), and \( \gamma = 500 \).  
In \textbf{SL} \(\to\) \textbf{SP} adaptation,  
when \( t \leq 200 \), the parameters are initialized as \( \lambda^g = \lambda^{cs} = \frac{1}{t + 1} \) and \( \gamma = 1 \).  
When \( t > 200 \), the parameters are updated to \( \lambda^g = 0.1 \), \( \lambda^{cs} = 0.01 \), and \( \gamma = 10 \).  

\noindent \textbf{Training Parameters.}\enspace  
The Stochastic Gradient Descent (SGD) optimizer is used for both pre-training and adaptation.  
In all experiments, the loss weights are set as \( w_{seg} = 1 \) and \( w_{dmc} = 1 \).  
In \textbf{SL} \(\to\) \textbf{SK} adaptation, training is conducted with a batch size of 8 and a learning rate of  \( 8e\text{-}4 \).  For~EMA update, the momentum factor is set to \( \lambda^N = 0.9 \), and the teacher network parameters are updated every \( \beta = 500 \) iterations.  
In \textbf{SL} \(\to\) \textbf{SP} adaptation, training is performed with a batch size of 2 and a learning rate of \( 1.4e\text{-}4 \). The~EMA update is configured with \( \lambda^N = 0.99 \), \( \beta = 1 \).

\subsection{Point Cloud UDA Semantic Segmentation}\enspace
The first two experiments (\tabref{semantickitti} and \tabref{semanticposs}) are designed to support that DPGLA outperforms the state-of-the-art in the task of point cloud UDA semantic segmentation.
\tabref{semantickitti} and \tabref{semanticposs} present the UDA semantic segmentation results on two synthetic-to-real tasks: SL\,\(\to\)\,SK and SL\,\(\to\)\,SP, respectively. 

In both tables, source denotes the model trained only on the labeled source domain (see \eqref{eq:source_loss}) without adaptation, which is highlighted with a yellow background in the first row.
The \textbf{best} results are highlighted in bold, while the \underline{second-best} results are underlined. 
The last row of the tables shows the \textit{improvement} over the source model.
In SL\,\(\to\)\,SK adaptation, DPGLA performs competitively and achieves 37.1\% mIoU, better than all competitors, surpassing second-best CoSMix by +3.9\%. Compared to the source model, it shows an improvement of +16.7\%.
In SL\,\(\to\)\,SP adaptation, DPGLA achieves 46.4\% mIoU, surpassing the second-best method, SALUDA, by +3.5\%, and improving +25.7\% over the source model.

Taking into account the results of both experiments, DPGLA shows a balanced overall improvement compared to other methods, rather than only a few classes standing out. 
This is made possible by our DPLF scheme, which ensures that the pseudo-labels involved in the training are more balanced and reliable.

\subsection{Ablation Studies}
The third experiment shows that each proposed module contributes to improving the model's performance, as summarized in \tabref{ablation}. We refer to each ablation study in the table by the letter in the first column.

To rigorously show the effectiveness of the modules, we first construct a baseline. It employs a self-training double-branch structure, LaserMix, local and global affine transformations, as well as standard uniform jitter, while the pseudo-label filtering uses a fixed threshold of 0.85~\cite{10.1007/978-3-031-19827-4_34}.
We then investigate the role of PG-DAP (\secref{sec:data_aug}), focusing on its three key components: DAS (A), DAJ (B), HAJ (C). 
Importantly, in our ablation study, DAJ and HAJ are not simply added in the baseline but replacing the standard uniform jitter to ensure a fair comparison.
The results show that each of the components of the PG-DAP is beneficial. Additionally, different jitter strategies are complementary, enhancing feature diversity and improving model adaptability (D and E).
Next, we explore the effect of the data mixing consistency loss (\secref{DMC}), which further refines the learned representations~(F). 
Finally, we focus on the DPLF scheme (\secref{DPLF}), a key component in our framework. We use DPLF to replace the fixed threshold 0.85. The results indicate that DPLF~(G) leads to a more beneficial training signal compared to a fixed threshold. 
The full model achieves the best overall performance, confirming that each component plays a vital role in improving adaptation robustness and effectiveness.

\section{CONCLUSIONS}
In this paper, we present DPGLA, a novel self-training-based approach for point cloud unsupervised domain adaptation semantic segmentation.
Our approach enables adaptation from synthetic to real 3D LiDAR point cloud within a Mean Teacher framework, exploiting a dynamic pseudo-label filtering scheme and a prior-guided data augmentation pipeline.
We evaluate our approach on different datasets and compared it with other existing methods.
The experiments demonstrate that DPGLA achieves state-of-the-art results.
Furthermore, we hope our findings offer valuable insights into point cloud UDA semantic segmentation and inspire the development of simpler and more effective approaches.

\balance
{
\bibliographystyle{IEEEtran}
\bibliography{IEEEabrv, references.bib}

\begin{thebibliography}{10}
\providecommand{\url}[1]{#1}
\csname url@samestyle\endcsname
\providecommand{\newblock}{\relax}
\providecommand{\bibinfo}[2]{#2}
\providecommand{\BIBentrySTDinterwordspacing}{\spaceskip=0pt\relax}
\providecommand{\BIBentryALTinterwordstretchfactor}{4}
\providecommand{\BIBentryALTinterwordspacing}{\spaceskip=\fontdimen2\font plus
\BIBentryALTinterwordstretchfactor\fontdimen3\font minus
  \fontdimen4\font\relax}
\providecommand{\BIBforeignlanguage}[2]{{%
\expandafter\ifx\csname l@#1\endcsname\relax
\typeout{** WARNING: IEEEtran.bst: No hyphenation pattern has been}%
\typeout{** loaded for the language `#1'. Using the pattern for}%
\typeout{** the default language instead.}%
\else
\language=\csname l@#1\endcsname
\fi
#2}}
\providecommand{\BIBdecl}{\relax}
\BIBdecl

\bibitem{8967762}
A.~Milioto, I.~Vizzo, J.~Behley, and C.~Stachniss, ``Rangenet ++: Fast and
  accurate lidar semantic segmentation,'' in \emph{IEEE/RSJ International
  Conference on Intelligent Robots and Systems (IROS)}, 2019.

\bibitem{9010727}
J.~Behley, M.~Garbade, A.~Milioto, J.~Quenzel, S.~Behnke, C.~Stachniss, and
  J.~Gall, ``Semantickitti: A dataset for semantic scene understanding of lidar
  sequences,'' in \emph{IEEE/CVF International Conference on Computer Vision
  (ICCV)}, 2019.

\bibitem{pmlr-v37-ganin15}
Y.~Ganin and V.~Lempitsky, ``Unsupervised domain adaptation by
  backpropagation,'' in \emph{Proceedings of the 32nd International Conference
  on Machine Learning (ICML)}, 2015.

\bibitem{10.1007/978-3-031-19827-4_34}
C.~Saltori, F.~Galasso, G.~Fiameni, N.~Sebe, E.~Ricci, and F.~Poiesi, ``Cosmix:
  Compositional semantic mix for domain adaptation in 3d lidar segmentation,''
  in \emph{Computer Vision -- ECCV}, 2022.

\bibitem{10160410}
L.~Kong, N.~Quader, and V.~E. Liong, ``Conda: Unsupervised domain adaptation
  for lidar segmentation via regularized domain concatenation,'' in \emph{IEEE
  International Conference on Robotics and Automation (ICRA)}, 2023.

\bibitem{10655228}
H.~Zhao, J.~Zhang, Z.~Chen, S.~Zhao, and D.~Tao, ``Unimix: Towards domain
  adaptive and generalizable lidar semantic segmentation in adverse weather,''
  in \emph{IEEE/CVF Conference on Computer Vision and Pattern Recognition
  (CVPR)}, 2024.

\bibitem{NEURIPS2022_475b85eb}
A.~Xiao, J.~Huang, D.~Guan, K.~Cui, S.~Lu, and L.~Shao, ``Polarmix: A general
  data augmentation technique for lidar point clouds,'' in \emph{Advances in
  Neural Information Processing Systems (NeurIPS)}, 2022.

\bibitem{10205234}
L.~Kong, J.~Ren, L.~Pan, and Z.~Liu, ``Lasermix for semi-supervised lidar
  semantic segmentation,'' in \emph{IEEE/CVF Conference on Computer Vision and
  Pattern Recognition (CVPR)}, 2023.

\bibitem{10.1007/978-3-031-72667-5_13}
L.~Li, H.~P.~H. Shum, and T.~P. Breckon, ``Rapid-seg: Range-aware pointwise
  distance distribution networks for 3d lidar segmentation,'' in \emph{Computer
  Vision -- ECCV}, 2025.

\bibitem{8793495}
B.~Wu, X.~Zhou, S.~Zhao, X.~Yue, and K.~Keutzer, ``Squeezesegv2: Improved model
  structure and unsupervised domain adaptation for road-object segmentation
  from a lidar point cloud,'' in \emph{International Conference on Robotics and
  Automation (ICRA)}, 2019.

\bibitem{8099499}
R.~Q. Charles, H.~Su, M.~Kaichun, and L.~J. Guibas, ``Pointnet: Deep learning
  on point sets for 3d classification and segmentation,'' in \emph{IEEE
  Conference on Computer Vision and Pattern Recognition (CVPR)}, 2017.

\bibitem{NIPS2017_d8bf84be}
C.~R. Qi, L.~Yi, H.~Su, and L.~J. Guibas, ``Pointnet++: Deep hierarchical
  feature learning on point sets in a metric space,'' in \emph{Advances in
  Neural Information Processing Systems}, 2017.

\bibitem{9010002}
H.~Thomas, C.~R. Qi, J.-E. Deschaud, B.~Marcotegui, F.~Goulette, and L.~Guibas,
  ``Kpconv: Flexible and deformable convolution for point clouds,'' in
  \emph{IEEE/CVF International Conference on Computer Vision (ICCV)}, 2019.

\bibitem{NEURIPS2022_d78ece66}
X.~Wu, Y.~Lao, L.~Jiang, X.~Liu, and H.~Zhao, ``Point transformer v2: Grouped
  vector attention and partition-based pooling,'' in \emph{Advances in Neural
  Information Processing Systems (NeurIPS)}, 2022.

\bibitem{8579059}
B.~Graham, M.~Engelcke, and L.~v.~d. Maaten, ``3d semantic segmentation with
  submanifold sparse convolutional networks,'' in \emph{IEEE/CVF Conference on
  Computer Vision and Pattern Recognition (CVPR)}, 2018.

\bibitem{9495168}
X.~Zhu, H.~Zhou, T.~Wang, F.~Hong, W.~Li, Y.~Ma, H.~Li, R.~Yang, and D.~Lin,
  ``Cylindrical and asymmetrical 3d convolution networks for lidar-based
  perception,'' \emph{IEEE Transactions on Pattern Analysis and Machine
  Intelligence (TPAMI)}, 2022.

\bibitem{10203638}
L.~Li, H.~P.~H. Shum, and T.~P. Breckon, ``Less is more: Reducing task and
  model complexity for 3d point cloud semantic segmentation,'' in
  \emph{IEEE/CVF Conference on Computer Vision and Pattern Recognition (CVPR)},
  2023.

\bibitem{10203552}
X.~Lai, Y.~Chen, F.~Lu, J.~Liu, and J.~Jia, ``Spherical transformer for
  lidar-based 3d recognition,'' in \emph{IEEE/CVF Conference on Computer Vision
  and Pattern Recognition (CVPR)}, 2023.

\bibitem{8953494}
C.~Choy, J.~Gwak, and S.~Savarese, ``4d spatio-temporal convnets: Minkowski
  convolutional neural networks,'' in \emph{IEEE/CVF Conference on Computer
  Vision and Pattern Recognition (CVPR)}, 2019.

\bibitem{NEURIPS2018_ab88b157}
M.~Long, Z.~Cao, J.~Wang, and M.~I. Jordan, ``Conditional adversarial domain
  adaptation,'' in \emph{Advances in Neural Information Processing Systems
  (NeurIPS)}, 2018.

\bibitem{ZELLINGER2019174}
W.~Zellinger, B.~A. Moser, T.~Grubinger, E.~Lughofer, T.~Natschläger, and
  S.~Saminger-Platz, ``Robust unsupervised domain adaptation for neural
  networks via moment alignment,'' \emph{Information Sciences}, 2019.

\bibitem{9080115}
J.~Li, E.~Chen, Z.~Ding, L.~Zhu, K.~Lu, and H.~T. Shen, ``Maximum density
  divergence for domain adaptation,'' \emph{IEEE Transactions on Pattern
  Analysis and Machine Intelligence (TPAMI)}, 2021.

\bibitem{8954439}
T.-H. Vu, H.~Jain, M.~Bucher, M.~Cord, and P.~Pérez, ``Advent: Adversarial
  entropy minimization for domain adaptation in semantic segmentation,'' in
  \emph{IEEE/CVF Conference on Computer Vision and Pattern Recognition (CVPR)},
  2019.

\bibitem{Liu_Han_Bai_Ge_Wang_Han_Li_You_Lu_2020}
X.~Liu, Y.~Han, S.~Bai, Y.~Ge, T.~Wang, X.~Han, S.~Li, J.~You, and J.~Lu,
  ``Importance-aware semantic segmentation in self-driving with discrete
  wasserstein training,'' \emph{Proceedings of the AAAI Conference on
  Artificial Intelligence}, 2020.

\bibitem{10.1145/3422622}
I.~Goodfellow, J.~Pouget-Abadie, M.~Mirza, B.~Xu, D.~Warde-Farley, S.~Ozair,
  A.~Courville, and Y.~Bengio, ``Generative adversarial networks,''
  \emph{Commun. ACM}, 2020.

\bibitem{9439889}
D.~Guan, J.~Huang, A.~Xiao, S.~Lu, and Y.~Cao, ``Uncertainty-aware unsupervised
  domain adaptation in object detection,'' \emph{IEEE Transactions on
  Multimedia}, 2022.

\bibitem{8953759}
R.~Gong, W.~Li, Y.~Chen, and L.~Van~Gool, ``Dlow: Domain flow for adaptation
  and generalization,'' in \emph{IEEE/CVF Conference on Computer Vision and
  Pattern Recognition (CVPR)}, 2019.

\bibitem{pmlr-v162-rangwani22a}
H.~Rangwani, S.~K. Aithal, M.~Mishra, A.~Jain, and V.~B. Radhakrishnan, ``A
  closer look at smoothness in domain adversarial training,'' in
  \emph{Proceedings of the 39th International Conference on Machine Learning
  (ICML)}, 2022.

\bibitem{8578878}
Y.-H. Tsai, W.-C. Hung, S.~Schulter, K.~Sohn, M.-H. Yang, and M.~Chandraker,
  ``Learning to adapt structured output space for semantic segmentation,'' in
  \emph{IEEE/CVF Conference on Computer Vision and Pattern Recognition (CVPR)},
  2018.

\bibitem{9372870}
Y.~Luo, P.~Liu, L.~Zheng, T.~Guan, J.~Yu, and Y.~Yang, ``Category-level
  adversarial adaptation for semantic segmentation using purified features,''
  \emph{IEEE Transactions on Pattern Analysis and Machine Intelligence
  (TPAMI)}, 2022.

\bibitem{9879466}
L.~Hoyer, D.~Dai, and L.~Van~Gool, ``Daformer: Improving network architectures
  and training strategies for domain-adaptive semantic segmentation,'' in
  \emph{IEEE/CVF Conference on Computer Vision and Pattern Recognition (CVPR)},
  2022.

\bibitem{9578759}
P.~Zhang, B.~Zhang, T.~Zhang, D.~Chen, Y.~Wang, and F.~Wen, ``Prototypical
  pseudo label denoising and target structure learning for domain adaptive
  semantic segmentation,'' in \emph{IEEE/CVF Conference on Computer Vision and
  Pattern Recognition (CVPR)}, 2021.

\bibitem{9889681}
Q.~Zhou, Z.~Feng, Q.~Gu, J.~Pang, G.~Cheng, X.~Lu, J.~Shi, and L.~Ma,
  ``Context-aware mixup for domain adaptive semantic segmentation,'' \emph{IEEE
  Transactions on Circuits and Systems for Video Technology}, 2023.

\bibitem{9010413}
Y.~Zou, Z.~Yu, X.~Liu, B.~V. K.~V. Kumar, and J.~Wang, ``Confidence regularized
  self-training,'' in \emph{IEEE/CVF International Conference on Computer
  Vision (ICCV)}, 2019.

\bibitem{9341508}
F.~Langer, A.~Milioto, A.~Haag, J.~Behley, and C.~Stachniss, ``Domain transfer
  for semantic segmentation of lidar data using deep neural networks,'' in
  \emph{IEEE/RSJ International Conference on Intelligent Robots and Systems
  (IROS)}, 2020.

\bibitem{9561255}
P.~Jiang and S.~Saripalli, ``Lidarnet: A boundary-aware domain adaptation model
  for point cloud semantic segmentation,'' in \emph{IEEE International
  Conference on Robotics and Automation (ICRA)}, 2021.

\bibitem{9578920}
L.~Yi, B.~Gong, and T.~Funkhouser, ``Complete \& label: A domain adaptation
  approach to semantic segmentation of lidar point clouds,'' in \emph{IEEE/CVF
  Conference on Computer Vision and Pattern Recognition (CVPR)}, 2021.

\bibitem{Zhao_Wang_Li_Wu_Gao_Xu_Darrell_Keutzer_2021}
S.~Zhao, Y.~Wang, B.~Li, B.~Wu, Y.~Gao, P.~Xu, T.~Darrell, and K.~Keutzer,
  ``epointda: An end-to-end simulation-to-real domain adaptation framework for
  lidar point cloud segmentation,'' \emph{Proceedings of the AAAI Conference on
  Artificial Intelligence}, 2021.

\bibitem{9945672}
Z.~Yuan, C.~Wen, M.~Cheng, Y.~Su, W.~Liu, S.~Yu, and C.~Wang, ``Category-level
  adversaries for outdoor lidar point clouds cross-domain semantic
  segmentation,'' \emph{IEEE Transactions on Intelligent Transportation
  Systems}, 2023.

\bibitem{Xiao_Huang_Guan_Zhan_Lu_2022}
A.~Xiao, J.~Huang, D.~Guan, F.~Zhan, and S.~Lu, ``Transfer learning from
  synthetic to real lidar point cloud for semantic segmentation,''
  \emph{Proceedings of the AAAI Conference on Artificial Intelligence}, 2022.

\bibitem{10007866}
Z.~Yuan, M.~Cheng, W.~Zeng, Y.~Su, W.~Liu, S.~Yu, and C.~Wang,
  ``Prototype-guided multitask adversarial network for cross-domain lidar point
  clouds semantic segmentation,'' \emph{IEEE Transactions on Geoscience and
  Remote Sensing}, 2023.

\bibitem{NIPS2017_68053af2}
A.~Tarvainen and H.~Valpola, ``Mean teachers are better role models:
  Weight-averaged consistency targets improve semi-supervised deep learning
  results,'' in \emph{Advances in Neural Information Processing Systems
  (NeurIPS)}, 2017.

\bibitem{10330760}
A.~Xiao, D.~Guan, X.~Zhang, and S.~Lu, ``Domain adaptive lidar point cloud
  segmentation with 3d spatial consistency,'' \emph{IEEE Transactions on
  Multimedia}, 2024.

\bibitem{10658318}
Z.~Yuan, W.~Zeng, Y.~Su, W.~Liu, M.~Cheng, Y.~Guo, and C.~Wang,
  ``Density-guided translator boosts synthetic-to-real unsupervised domain
  adaptive segmentation of 3d point clouds,'' in \emph{IEEE/CVF Conference on
  Computer Vision and Pattern Recognition (CVPR)}, 2024.

\bibitem{10550726}
B.~Michele, A.~Boulch, G.~Puy, T.-H. Vu, R.~Marlet, and N.~Courty, ``Saluda:
  Surface-based automotive lidar unsupervised domain adaptation,'' in
  \emph{2024 International Conference on 3D Vision (3DV)}, 2024.

\bibitem{7785132}
F.~Milletari, N.~Navab, and S.-A. Ahmadi, ``V-net: Fully convolutional neural
  networks for volumetric medical image segmentation,'' in \emph{Fourth
  International Conference on 3D Vision (3DV)}, 2016.

\bibitem{NEURIPS2021_c1fea270}
H.~Liu, J.~Wang, and M.~Long, ``Cycle self-training for domain adaptation,'' in
  \emph{Advances in Neural Information Processing Systems (NeurIPS)}, 2021.

\bibitem{10839256}
J.-i. Park, S.~Jo, H.-T. Seo, and J.~Park, ``Lidar denoising methods in adverse
  environments: A review,'' \emph{IEEE Sensors Journal}, 2025.

\bibitem{8099799}
E.~Tzeng, J.~Hoffman, K.~Saenko, and T.~Darrell, ``Adversarial discriminative
  domain adaptation,'' in \emph{IEEE Conference on Computer Vision and Pattern
  Recognition (CVPR)}, 2017.

\bibitem{9304596}
Y.~Pan, B.~Gao, J.~Mei, S.~Geng, C.~Li, and H.~Zhao, ``Semanticposs: A point
  cloud dataset with large quantity of dynamic instances,'' in \emph{IEEE
  Intelligent Vehicles Symposium (IV)}, 2020.

\bibitem{10.1007/978-3-319-50835-1_22}
M.~A. Rahman and Y.~Wang, ``Optimizing intersection-over-union in deep neural
  networks for image segmentation,'' in \emph{Advances in Visual Computing},
  2016.

\end{thebibliography}
}
\end{document}